\documentclass[runningheads]{llncs}

\usepackage{eccv}
\usepackage{eccvabbrv}
\usepackage{graphicx}
\usepackage{booktabs}
\usepackage{tikz}
\usetikzlibrary{arrows.meta,positioning}
\usepackage[accsupp]{axessibility}
\usepackage{hyperref}
\hypersetup{
  pdftitle={Aggregate, Don't Adapt: Subject-Level Posterior Aggregation and Transductive
            Calibration for Cross-Site Parkinsonian Gait Severity},
  pdfauthor={Junlong Shen},
  pdfsubject={MoCha 2026 Benchmark and Challenge on Parkinsonian Gait -- challenge report},
  pdfkeywords={Parkinson's disease; gait analysis; cross-site generalization;
               transductive inference; clinical motion analysis}}

\newcommand{\TEAMNAME}{JLShen}
\newcommand{\AUTHORNAME}{Junlong Shen}
\newcommand{\AUTHORAFFIL}{University of Alberta, Edmonton, Canada}
\newcommand{\AUTHOREMAIL}{junlong6@ualberta.ca}

\begin{document}

\title{Aggregate, Don't Adapt: Subject-Level Posterior\\Aggregation and Transductive Calibration for\\Cross-Site Parkinsonian Gait Severity}
\titlerunning{Aggregate, Don't Adapt}

\author{\AUTHORNAME}
\authorrunning{\AUTHORNAME}
\institute{\AUTHORAFFIL\\ \email{\AUTHOREMAIL}}

\maketitle

\begin{abstract}
We describe the winning entry to the MoCha 2026 Benchmark and Challenge on Parkinsonian Gait,
which predicts MDS-UPDRS gait severity from canonicalized SMPL motion recorded at clinical sites
unseen during training. The system reaches \textbf{0.69447} macro-F1 on the hidden test and
ranked \textbf{first of 58 entries}, ahead of the runner-up at 0.5807 and the organizers'
baseline at 0.4289, on a frozen public motion encoder with a single $4\times512$ linear layer.
Nearly all of the margin comes from three stages usually treated as bookkeeping: reproducing the
reference benchmark's exact head recipe, averaging per-walk posteriors within the subject
grouping the organizers ship, and a label-free transductive calibration of the feature mean and
the decision operating point. Fine-tuning the encoder lost in four distinct forms, and ten
alternative encoders were worse. Every ablation number is a paid read on the hidden test,
because our own leave-two-cohort-out cross-validation proved anti-correlated with the deciding
score over eleven configurations. We give the negative record in full, and identify our largest
gain, subject-level aggregation, as the binding ceiling on this benchmark.

\end{abstract}

\section{Introduction}

Predicting parkinsonian gait severity at a clinical site never seen in training is a
cross-site generalization problem, and on this benchmark the decisive levers turn out not to
live in the motion representation. The MoCha 2026 challenge~\cite{mocha2026} asks for the MDS-UPDRS gait severity
class~\cite{updrs} of a walking sequence given as canonicalized SMPL motion~\cite{smpl}.
Training data is the CARE-PD benchmark~\cite{adeli2025carepd}; the hidden test comes from
clinical sites absent from training. Entries are ranked by macro-F1. Submissions are code:
participants upload a \texttt{predict} function with model weights, and the organizers run it
on hidden motion files.\footnote{This is the arXiv version of the challenge report submitted to
the MoCha 2026 organizers for prize verification. It is not part of the workshop proceedings.}

\begin{figure}[t]
\centering
\begin{tikzpicture}[
  font=\scriptsize,
  box/.style={draw, rounded corners=1.2pt, minimum height=7.5mm, minimum width=20mm,
              align=center, inner sep=1.5pt},
  rep/.style={box, fill=black!4, draw=black!45},
  win/.style={box, fill=blue!10, draw=blue!60, thick},
  ar/.style={-{Latex[length=1.3mm]}, draw=black!55},
]
\node[rep] (a1) {canonicalized\\SMPL walk};
\node[rep, right=2.6mm of a1] (a2) {joints, side view\\81-frame clips};
\node[rep, right=2.6mm of a2] (a3) {\textbf{frozen}\\MotionAGFormer-S};
\node[win, right=2.6mm of a3] (a4) {transductive\\centering};
\node[win, right=2.6mm of a4] (a5) {$4{\times}512$ linear\\head, 14\,KB};
\foreach \i/\j in {a1/a2, a2/a3, a3/a4, a4/a5} \draw[ar] (\i) -- (\j);

\node[rep, below=6.5mm of a1.south west, anchor=north west] (b1) {per-walk\\posteriors};
\node[win, right=2.6mm of b1] (b2) {\textbf{subject mean} over\\\texttt{data[sid][wid]}};
\node[win, right=2.6mm of b2] (b3) {subject-kNN\\pooling};
\node[win, right=2.6mm of b3] (b4) {$q$-divisor\\operating point};
\node[rep, right=2.6mm of b4] (b5) {one label\\per subject};
\foreach \i/\j in {b1/b2, b2/b3, b3/b4, b4/b5} \draw[ar] (\i) -- (\j);
\draw[ar] (a5.south) -- ++(0,-2.2mm) -| (b1.north);

\begin{scope}[shift={(0,-2.55)}]
  \def\sc{36}   
  \foreach \y/\d/\lab in {%
     0/0.143/{subject-mean aggregation},
    -0.36/0.057/{benchmark-exact head recipe},
    -0.72/0.017/{transductive centering $+$ operating point},
    -1.08/0.0075/{subject-kNN pooling}} {
    \node[anchor=east, font=\scriptsize] at (4.55,\y) {\lab};
    \fill[blue!55, rounded corners=0.6pt] (4.7,\y-0.10) rectangle (4.7+\d*\sc,\y+0.10);
    \node[anchor=west, font=\scriptsize] at (4.7+\d*\sc+0.08,\y) {$+\d$};
  }
  \node[anchor=east, font=\scriptsize, text=black!60] at (4.55,-1.50)
        {4 fine-tuning forms, 10 alternative encoders};
  \node[anchor=west, font=\scriptsize, text=black!60] at (4.7,-1.50) {all lost};
\end{scope}
\end{tikzpicture}
\caption{\textbf{The winning margin came from reading the input format, not from the motion
representation.} The encoder is public and frozen; the only trained parameters are a 14\,KB
linear layer (top). Averaging per-walk posteriors within the subject grouping that the
organizers ship inside the input is worth $+0.143$ macro-F1, more than every representation
change we made combined (bottom); every bar is a paid read on the hidden test. That same
aggregation is also this benchmark's ceiling (\cref{sec:limits}).}
\label{fig:teaser}
\end{figure}
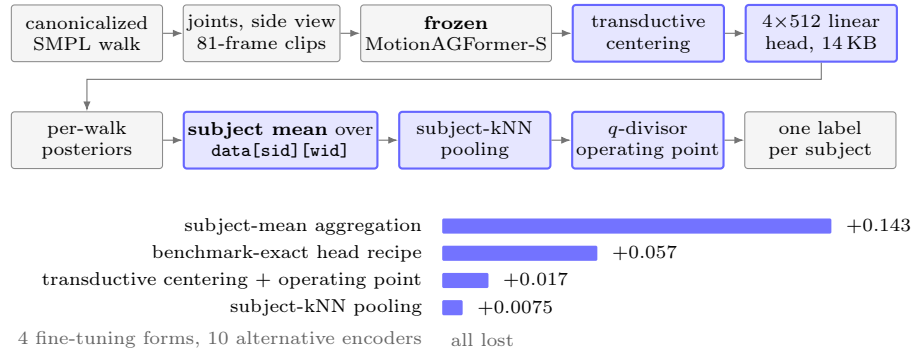

Our starting hypothesis was the usual one. We assumed the gap to a useful system was a
representation gap, and that a better or better-adapted motion encoder would close it. That
hypothesis failed in every form we tested. What moved the score instead was aggregating
predictions over the subject grouping that ships inside the input format, and reproducing the
reference benchmark's training recipe exactly rather than approximately. Two label-free
transductive corrections that read only the unlabeled test distribution account for the
remainder. \cref{fig:teaser} puts the four measured contributions on one axis.

\paragraph{Contributions.}
\begin{enumerate}
\item A system that reaches \textbf{0.69447} macro-F1 on the MoCha 2026 hidden test, rank 1 of
      58, $+0.114$ over the runner-up and $+0.266$ over the released baseline, from
      \textbf{14\,KB} of trained parameters on a frozen public encoder.
\item A component ablation in which every number is a paid read on the deciding test set
      (\cref{sec:results}), including the finding that subject-level posterior aggregation is
      worth more ($+0.143$) than every model change we made combined.
\item A measurement result that generalizes beyond this challenge: our leave-two-cohort-out
      cross-validation was \emph{anti}-correlated with the hidden test (\cref{sec:cv}).
\item The full negative record (\cref{sec:negative}), and the ceiling our own best stage imposes
      (\cref{sec:limits}).
\end{enumerate}

\section{Task, data and metric}
\label{sec:task}

\subsection{What goes in and what comes out}

The unit of prediction is a \emph{walk}, and walks arrive already grouped by subject. The
evaluation server calls \texttt{predict(data)} once with the entire hidden test.
\texttt{data[subject\_id][walk\_id]} holds one canonicalized SMPL sequence: an axis-angle pose
array of shape $(T,72)$, a global translation of shape $(T,3)$, a shape vector zeroed for
privacy, and the original capture frame rate. The return value is
\texttt{predictions[subject\_id][walk\_id]}, an integer severity class in $\{0,1,2,3\}$, one per
walk. Missing or invalid predictions count as wrong.

\begin{figure}[t]
\centering
\includegraphics[width=\textwidth]{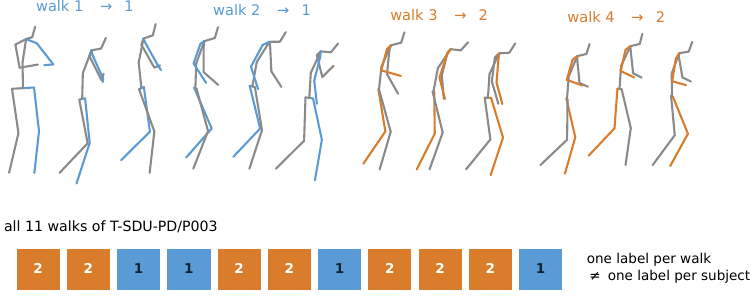}
\caption{\textbf{Severity is labeled per walk, and a subject's walks do not have to agree.}
Four walks of one T-SDU-PD subject, drawn from the released canonicalized SMPL motion through
forward kinematics and projected to the side view the encoder reads; three poses are shown per
walk. Two carry ground-truth class 1 and two carry class 2. The strip gives all eleven walks of
that subject. Averaging over this grouping is what buys $+0.143$ macro-F1
(\cref{sec:results}); the residual disagreement inside it is what caps the system
(\cref{sec:limits}).}
\label{fig:task}
\end{figure}

That two-level dictionary is the single most important fact about the task. The subject
identity is anonymized, but the \emph{grouping} is given, at test time, for free. \cref{fig:task}
shows what one subject's walks look like and how their labels behave.

\subsection{Label space}

Three properties of the label space drive every design decision later, and \cref{tab:labelspace}
holds all three. Training uses CARE-PD~\cite{adeli2025carepd}, a multi-site anonymized clinical
dataset released as nine cohorts of canonicalized SMPL walks, four of which carry the
\texttt{UPDRS\_GAIT} annotation this task predicts. Severity 3 is rare in those four, at 44 walks
in 2952, and absent from two cohorts entirely. Cohort class frequencies differ sharply, with
class 2 ranging from 0.15 in PD-GaM to 0.44 in T-SDU-PD. And 62 of 110 subjects carry more than
one label across their own walks.

\begin{table}[t]
\centering
\caption{The four UPDRS-labeled CARE-PD cohorts, counted from the released pickles. Columns
0--3 count walks per severity class, and \emph{mixed} counts subjects whose own walks do not all
carry the same label. Class 3 is rare and missing from two cohorts, and more than half of all
subjects are mixed.}
\label{tab:labelspace}
\small
\begin{tabular}{lrrrrrrr}
\toprule
cohort & subjects & walks & 0 & 1 & 2 & 3 & mixed \\
\midrule
3DGait & 43 & 90 & 24 & 42 & 14 & 10 & 16 \\
PD-GaM & 30 & 1700 & 783 & 635 & 248 & 34 & 29 \\
BMCLab & 23 & 781 & 341 & 276 & 164 & 0 & 8 \\
T-SDU-PD & 14 & 381 & 96 & 118 & 167 & 0 & 9 \\
\midrule
all four & 110 & 2952 & 1244 & 1071 & 593 & 44 & 62 \\
\bottomrule
\end{tabular}

\end{table}

The remaining five cohorts (DNE~\cite{hoang2024smartphone,hoang2022towards},
E-LC~\cite{mckay2019freezing,kwon2023explainable},
KUL-DT-T~\cite{spildooren2010freezing,filtjens2022automated}, T-LTC and
T-SDU~\cite{adeli2025carepd}) ship without severity labels. We used them only as unlabeled data
in a semi-supervised arm that we report as a negative result in \cref{sec:negative}. No cohort
outside CARE-PD entered the system.

\subsection{Metric}

The ranking metric has one degree of freedom that no provided artifact fixes, and it is worth a
quarter of the score. Entries are ranked by macro-F1, with macro-precision, macro-recall,
accuracy and quadratic weighted kappa (QWK) reported alongside. The challenge pages state the
primary metric as ``Macro F1'' and no further. They do not say which class set the average runs
over, and the starting kit ships a baseline and a submission API but not the scoring program, so
the question cannot be settled by reading provided code. It matters because severity 3 is rare
enough that a system can miss it entirely, and dividing by three rather than four changes the
score by roughly a quarter of one class. We resolved it from the leaderboard, by submitting a constant predictor and reading
back the score it earned. The server divides by four, so an unpredicted class costs a full
quarter of the metric. Every number in this report is that four-class macro-F1.

\section{Method}

The system is a frozen encoder with a linear probe and three post-hoc stages, and no encoder is
fine-tuned anywhere in it.

\begin{enumerate}
\item \textbf{Geometry.} SMPL pose and translation go through forward kinematics to 17 3D
      joints. Those joints are decimated to approximately 30\,fps by striding on the
      \texttt{fps} field shipped with each sequence, projected to a single side view, cut into
      81-frame clips, and normalized by a per-clip crop-scale.
\item \textbf{Frozen features.} A MotionAGFormer-S encoder~\cite{motionagformer} pretrained on
      Human3.6M~\cite{h36m}, used unmodified, produces a per-clip $(81,17,512)$ representation.
      We average over valid frames, then over clips, then over the 17 joint tokens, giving one
      512-d vector per walk. Joint-mean pooling beat keeping the per-joint 8704-d
      representation, which is why the head is small.
\item \textbf{Standardization.} Features are z-scored with the \texttt{feat\_mean} and
      \texttt{feat\_std} stored inside the trained head.
\item \textbf{Transductive centering.} We subtract a shrunken estimate of the \emph{test}
      feature mean, $X \leftarrow X - c\,\bar{X}_{\text{test}}$ with $c=0.90$. This is the only
      distribution-alignment operation that transferred; \cref{sec:negative} lists the eight
      that did not.
\item \textbf{Linear head.} A single $4\times512$ linear layer trained with focal
      loss~\cite{focal} ($\alpha=1$, $\gamma=1$) and AdamW, producing per-walk posteriors.
\item \textbf{Subject-mean aggregation.} Per-walk posteriors are averaged within each subject id
      of the released \texttt{data[subject\_id][walk\_id]} grouping, at $\lambda = 1.0$, so every
      walk of a subject receives the same label. The aggregator is the uniform arithmetic mean.
\item \textbf{Subject-kNN posterior pooling.} Each subject's posterior is blended toward its
      $k=5$ nearest \emph{subjects} by cosine similarity on subject-mean embeddings, with weight
      $\lambda_{\text{nbr}} = 0.40$.
\item \textbf{Operating point.} A $q$-divisor logit adjustment~\cite{logitadjust},
      $P' \propto P / q^{\tau}$ with $\tau = 0.50$, where $q$ is the model's own \emph{predicted}
      class marginal on the test set, followed by \texttt{argmax}.
\end{enumerate}

\paragraph{Two properties we want stated explicitly.}
The system is \emph{transductive but label-free}. \texttt{predict()} receives the entire hidden
test at once, and stages 4 and 8 read only unlabeled test statistics: its feature mean and its
own predicted class marginal. No hidden-test label is used, inferred or reconstructed at any
point, no hidden-test data was downloaded, scraped or exported from the evaluation server, and
no attempt was made to re-identify any subject. The only grouping used is the anonymized subject
id supplied in the input. The system also \emph{exploits released structure, not leakage}: that
grouping is part of the input format the organizers ship, and aggregating over it is the single
largest contribution here. Training used the CARE-PD cohorts and a public pretrained encoder and
nothing else.

\paragraph{Training.} Feature extraction runs the frozen encoder over the four UPDRS-labeled
CARE-PD cohorts (110 subjects, 2952 walks) under three augmentations that transferred, namely
spatial jitter, temporal speed-warp and frame dropout, giving 44280 augmented rows. The head is
then trained in two phases following the reference benchmark protocol: the epoch count is tuned
on a 15\% within-train split by macro-F1, then the head is refit on 100\% of the labeled data
for that epoch count. Total training cost is under one GPU-hour, dominated by feature
extraction. Inference needs only \texttt{torch} and \texttt{numpy}, since the SMPL forward
kinematics and the joint regressor are vendored into the submission. It is far inside the
3600\,s evaluation limit: re-runs of the delivered artifact took 22--41\,s for a 284-walk probe
including model load on one partition of an H100 GPU, against a 372-walk real test.

\section{How we measured}
\label{sec:cv}

\begin{figure}[t]
\centering
\includegraphics[width=0.62\textwidth]{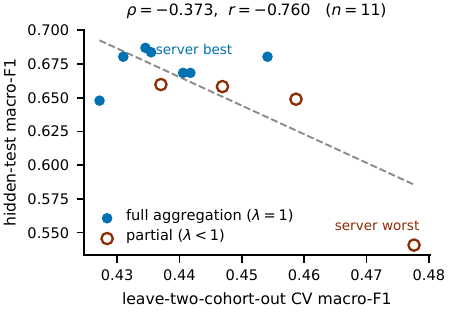}
\caption{\textbf{Our local cross-validation ranked configurations backwards.} Each point is one
of the eleven configurations for which we hold both a leave-two-cohort-out score and a paid
hidden-test read. Filled points use full subject aggregation, open points use less. The
inversion is concentrated on the aggregation axis, which is also the axis worth the most.}
\label{fig:cvserver}
\end{figure}

Our cross-validation was anti-correlated with the score that decides the competition, and
finding that out was the most useful measurement we made. Partway through the challenge we ran
a calibration we should have run at the start. We replayed the eleven configurations for which
we held paid hidden-test reads through our own local cross-validation, a nested
leave-two-cohort-out protocol over the four labeled cohorts, and rank-correlated the two.
\cref{fig:cvserver} shows the result: Spearman $\rho = -0.373$ and Pearson $r = -0.760$. The
server's best configuration scored close to the lowest locally, and the server's worst scored
the highest.

The inversion is worst exactly where it costs most. Cross-validation preferred no subject
aggregation by $-0.042$, while the hidden test preferred full aggregation by $+0.143$. This is
a property of the estimator rather than a bug in it. Shuffling the training labels collapses the
same estimator from 0.4416 to 0.2234, so it does measure real signal. It simply measures a
quantity whose ordering is inverted with respect to cross-site transfer for these levers. Our
reading is that leave-two-cohort-out folds reward configurations that fit the idiosyncrasies of
whichever cohorts remain, and subject aggregation deliberately discards exactly that kind of
within-cohort detail.

We therefore retired cross-validation as a selector for every decision, aggregation and
operating-point lever, and selected on hidden-test reads instead. That is defensible here
because the challenge is a single-phase evaluation on a fixed test set with a deterministic
scorer. There is no private re-split behind the visible board, so maximizing over reads is exact
optimization of the announced objective rather than selection on noise. Every number in
\cref{sec:results} is consequently a paid read on the hidden test.

Two practical consequences shaped the rest of the run. First, each submission spends one of
three daily evaluation slots, which is why we call every hidden-test measurement a \emph{paid
read}. Reads are scarce, so we screened candidates offline for \emph{distinctness}, running
each deployable artifact on a local probe and diffing the returned label dictionaries.
Configurations that emit identical predictions are zero-information submissions. That screen
killed a $k$-sweep, an entire evidence-weighting family, and four of six operating-point
re-tune cells for zero submissions. Second, we re-derived the evaluation set's \emph{size} from
the leaderboard's own \texttt{accuracy} column, since every reported accuracy is an exact
rational $k/N$: 40 readings of ours and all ten visible competitors' land on $N=372$. This is
arithmetic on publicly displayed leaderboard numbers, it involves no test datum or label, and it
conferred no scoring advantage. It did correct an internal premise that had wrongly closed a
lever family. A premise-based kill dies with its premise.

\section{Results}
\label{sec:results}

\subsection{Final standing}

The field converged on the reference recipe and we did not, which is where the 0.114 gap comes
from. In \cref{tab:board} the entire visible board except our entry sits inside the published
CARE-PD cross-site band of roughly 0.52--0.55 macro-F1 for a single frozen encoder with a linear
probe~\cite{adeli2025carepd}. The rivals sit about 0.11 below us on \emph{accuracy} while sharing our
precision-leaning profile, so the difference is base separability rather than operating-point
tuning.

\begin{table}[t]
\centering
\caption{Final standing on the MoCha 2026 hidden test. Every column is the organizers' scorer on
the hidden test; our ranked entry is Codabench submission 882979. Leaderboard 18564, phase
27428, read at the close of the phase on 2026-08-16, when the board carried 58 entries. Boards
stay editable after a phase closes: a re-read on 2026-08-20 listed 50 entries with this ordering
unchanged.}
\label{tab:board}
\small
\begin{tabular}{clccccc}
\toprule
\# & team & macro-F1 & macro-P & macro-R & accuracy & QWK \\
\midrule
1 & \textbf{\TEAMNAME{} (ours)} & \textbf{0.69447} & 0.72347 & 0.67659 & 0.65591 & 0.60007 \\
2 & brady\_kinesia    & 0.58070 & 0.64650 & 0.55449 & 0.53495 & 0.40674 \\
3 & unist\_visionlab  & 0.57358 & 0.56464 & 0.58659 & 0.54301 & 0.42932 \\
4 & Nottingham\_RVCE  & 0.55577 & 0.59830 & 0.53448 & 0.53763 & 0.42050 \\
5 & tuananh1007       & 0.54907 & 0.59898 & 0.52355 & 0.48925 & 0.43321 \\
6 & anhnamxtanh       & 0.54097 & 0.58588 & 0.51573 & 0.50806 & 0.40438 \\
\midrule
\multicolumn{2}{l}{\emph{organizers' released baseline}} & 0.42890 & --- & --- & --- & --- \\
\bottomrule
\end{tabular}
\end{table}

\subsection{Which stage bought the margin}

One row of \cref{tab:ladder} is worth more than all the others together, and the four questions
below take the ladder in order of what each stage settles.

\begin{table}[t]
\centering
\caption{Component ladder. Every row is a paid read of the organizers' scorer on the hidden test,
never a local estimate. Rows were measured at different points in the run, so each $\Delta$ is
against the row above it on the pipeline as it then stood, not a strictly nested leave-one-out
ablation.}
\label{tab:ladder}
\small
\begin{tabular}{lcc}
\toprule
stage & macro-F1 & $\Delta$ \\
\midrule
organizers' released baseline bundle                              & 0.4289 & --- \\
frozen MotionAGFormer-S + plain linear probe                      & 0.467  & $+0.038$ \\
\quad + benchmark-exact head recipe (focal loss, AdamW, z-score)  & 0.524  & $+0.057$ \\
\quad + transductive centering \& $q$-divisor operating point     & 0.5407 & $+0.017$ \\
\quad + \textbf{subject-mean posterior aggregation} ($\lambda=1$) & 0.68371& $\mathbf{+0.143}$ \\
\quad + centering shrinkage retuned $c\,{=}\,0.8 \rightarrow 0.9$ & 0.68699& $+0.0033$ \\
\quad + subject-kNN posterior pooling ($k{=}5$, $\lambda_{\text{nbr}}{=}0.40$)
                                                                  & \textbf{0.69447} & $+0.0075$ \\
\bottomrule
\end{tabular}
\end{table}

\paragraph{Does the architecture or the training recipe carry the reference system? The recipe,
worth $+0.057$.} Our first probe on frozen MotionAGFormer-S features was a plain
logistic-regression head with cross-entropy loss, and it plateaued at 0.467 for weeks in a way
that read like a task ceiling. It was not a ceiling. Reproducing the CARE-PD benchmark's
\emph{exact} head recipe moved the same features to 0.524: focal loss with $\alpha=1,\gamma=1$,
AdamW, and z-scored features, taken from the benchmark's vendored winning-configuration JSON
files rather than from prose. The architecture was never the problem. When reproducing a
reference system, the training recipe is part of the system.

\paragraph{How much is the released subject grouping worth? More than every model change
combined, at $+0.143$.} The input format groups walks by subject, and the hidden test is no
exception. Averaging per-walk posteriors within a subject and emitting one label per subject
beats every representation change we made put together. The sweep is monotone to the boundary:
$\lambda = 0.00 \rightarrow 0.5407$, $0.75 \rightarrow 0.6489$, $0.90 \rightarrow 0.65828$,
$1.00 \rightarrow 0.68371$. The choice of aggregator matters too. The uniform arithmetic mean
beat confidence-weighted pooling (0.66416), trimmed-25\% pooling (0.65837) and logit-space
geometric pooling (0.60777); geometric pooling sharpens posteriors and cripples the downstream
operating point. On this benchmark, aggregating over a grouping the organizers \emph{shipped} is
a first-class modeling decision rather than data plumbing.

\paragraph{Can evidence be borrowed across subjects as well as within one? Yes, but only
locally, and only for $+0.0075$.} Blending a subject's posterior toward its $k$ nearest subjects
has a sharp interior optimum: $\lambda_{\text{nbr}}$ of $0 \rightarrow 0.68699$, $0.40
\rightarrow \mathbf{0.69447}$, $0.45 \rightarrow 0.67256$, $0.70 \rightarrow 0.62769$, $1.00
\rightarrow 0.4012$. Its own limit confirms the mechanism. Blending toward the global test-mean
posterior, the $k = \infty$ case, scores 0.67517, \emph{below} base. Local pooling denoises and
global pooling washes out genuine between-subject severity differences. The value of $k$ itself
is flat, with $k \in \{3,8,12\}$ byte-identical to $k=5$ at $\lambda_{\text{nbr}} = 0.40$.

\paragraph{Are the two continuous knobs tuned or merely lucky? Both are saturated and
piecewise-constant.} Centering shrinkage gives $c\,0.60 \rightarrow 0.68042$, $c\,0.70 = c\,0.80
\rightarrow 0.68371$, and $c\,0.90 = c\,0.95 = c\,1.00 \rightarrow 0.68699$. The operating point
gives $\tau\,0.40 = \tau\,0.45 \rightarrow 0.66843$, $\tau\,0.50 = \tau\,0.52 \rightarrow
0.68371$, $\tau\,0.55 \rightarrow 0.68042$, and $\tau\,0.60 \rightarrow 0.64783$. Both were
re-swept after the pooling stage was added, since pooling changes exactly the posterior
peakedness that the operating point reads, and both held.

\section{What did not work}
\label{sec:negative}

For a challenge report this is the more informative half, and we give it in full.
\cref{tab:negative} lists each family we closed by direct measurement rather than by intuition.

\begin{table}[t]
\centering
\caption{Rejected families, each closed by measurement. ``Server'' marks a paid read of the
organizers' scorer on the hidden test; the remaining entries are local cross-validation, which
\cref{sec:cv} shows is a weak selector on decision and aggregation axes but a usable one for
gross representation failure.}
\label{tab:negative}
\small
\setlength{\tabcolsep}{4pt}
\begin{tabular}{p{3.0cm}p{8.5cm}}
\toprule
family & outcome \\
\midrule
Encoder fine-tuning &
Dead in four forms: unsupervised 2D$\rightarrow$3D pretext fine-tuning (server 0.3605, far below
frozen), anti-site-latch low-rank adaptation (LoRA) with gradient reversal and a bottleneck, supervised
LoRA, and
contrastive self-supervision with a gradient reversal layer. \\
Alternative encoders &
All below MotionAGFormer-S: MixSTE~\cite{mixste} (server 0.4081, including in its own native
normalization pipeline), MotionAGFormer-B/L, MotionBERT~\cite{motionbert}, PoseFormerV2,
MotionCLIP, TMR, an NTU-pretrained skeleton transformer, MoMask, a probe on the Riemannian
manifold of symmetric positive-definite matrices, and a time-series foundation model. \\
Alignment beyond the first moment &
Only the mean transfers. CORAL~\cite{coral} and diagonal-std alignment, median centering, PCA
denoising, iterative nullspace projection, transductive batch normalization, quantile-CDF
matching, and per-site clustered centering all lost. The last lost even with \textbf{oracle site
labels}, so it is not a clustering-quality problem. \\
Decision rules and objectives &
Ordinal and CORAL-style ordinal heads (server 0.4558, $-0.23$, bracketed over $\tau$ so not a
mistuning artifact), threshold tuning, linear discriminant analysis, Saerens EM label
shift~\cite{saerens}, class-balanced centering, prototype and nearest-class-mean rules, a soft
macro-F1 surrogate objective, the invariance objectives V-REx, GroupDRO~\cite{groupdro} and IRM,
and training-time de-confidencing. A
parameter-free joint macro-F1 decoder returned predictions byte-identical to \texttt{argmax} on
the server, so \textbf{argmax is already F1-optimal here}, and the recall gap is separability
rather than decision-rule miscalibration. Per-class decision scaling also lost in both
directions: suppressing class 3 gave 0.64882 and boosting class 2 gave 0.5715. \\
Ensembles &
Server-negative in every form: diverse-encoder blends, MotionAGFormer $\oplus$ MixSTE (locally
positive, server collapse to 0.3720), fine-tuned $\oplus$ frozen, and two-view side and back
(cross-validation $+0.032$ inverted to $-0.042$ on the server). See the note below. \\
Transductive inference beyond centering &
Graph label propagation~\cite{lgc}, Nystr{\"o}m-RBF and metric learning, density-ratio
importance weighting, FixMatch~\cite{fixmatch} over the five unlabeled cohorts, Deep Feature
Reweighting~\cite{dfr}, attention pooling over clips, bag-of-windows pooling, and self-training
test-time adaptation~\cite{tent}. The last was an exact no-op, changing 0 of 284 probe items. \\
Hand-crafted clinical gait features &
Cadence, stride, asymmetry and variability features are \textbf{96--98\% cohort-predictable}:
they act as site detectors on this data, and a random-forest late blend improved the pooled mean
while hurting held-out cohorts. A useful by-product is that $k$-means on those same features
recovers the cohorts at $\approx 1.00$ purity, so they are an excellent \emph{site} signal and a
poor \emph{severity} signal. \\
\bottomrule
\end{tabular}
\end{table}

\paragraph{Does an ensemble still help once the pipeline already averages? Not measurably.}
Eight-head same-recipe averaging cost $-0.04$ on the pre-aggregation base, at 0.5007 against
0.5407. We re-measured it on the final system, since a negative result is only valid for the
base it was measured on. Two-head averaging of the two strongest heads gives 0.69106, five-head
averaging also gives 0.69106, and two-head averaging with the most \emph{decorrelated} head
gives 0.68764, against 0.69447 for the single head. On the aggregated base, ensembling is
roughly neutral at $-0.003$ rather than a collapse: subject-mean and kNN pooling already perform
the variance reduction an ensemble would supply. A single-head no-op control through the same
code path reproduced the deployed predictions exactly, so these numbers isolate the averaging.
Selecting a voter by \emph{solo} score beat selecting it by decorrelation, which is the opposite
of the usual guidance.

\section{Where the ceiling is}
\label{sec:limits}

\subsection{Our best stage is also our ceiling}

\begin{figure}[t]
\centering
\includegraphics[width=\textwidth]{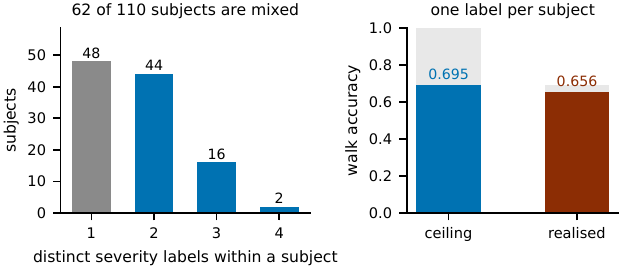}
\caption{\textbf{One label per subject is worth $+0.143$, and it is also a hard bound.} Left:
most subjects in the labeled cohorts carry more than one severity label across their own walks.
Right: predicting one label per subject can reach at most 69.5\% walk accuracy on those cohorts,
and the deployed system realizes 65.6\% on the hidden test. Both panels count released labels
only.}
\label{fig:ceiling}
\end{figure}

The stage that bought the margin is the same stage that now bounds it. On the labeled cohorts,
62 of 110 subjects carry \emph{mixed} walk labels: 44 span two severity classes, 16 span three,
and 2 span all four (\cref{fig:ceiling}). Forcing one label per subject therefore makes 901 of 2952 walks
unreachable, or 30.5\%, a hard walk-accuracy ceiling of 69.5\%. We realize 65.59\%. The
$\lambda = 1.0$ family is thus about 95\% saturated, and every lever inside it fights over a
handful of walks while the aggregation stage discards far more. Every attempt we made to
re-admit per-walk evidence on top of aggregation lost. Recovering within-subject variation
without giving up the variance reduction that aggregation buys is the open problem this
benchmark now poses.

\subsection{The head is unstable to its random initialization}

Part of our margin is a favorable draw, and the paper would be misread without that stated.
Retraining the identical recipe with different seeds gives hidden-test macro-F1 of 0.694, 0.681,
0.652, 0.606 and 0.558 across five draws, a spread of 0.14 on a $4\times512$ linear probe fit to
110 subjects. The shipped head is the best of those draws, which is also why we release the
exact trained head rather than a training script alone. Seed ensembling does not remove the
variance, as \cref{sec:negative} shows. A difference of $\pm0.02$ between two systems on this
benchmark should not be read as a method effect.

\subsection{A single operating point is wrong for every site at once}

Per-cohort severity distributions differ sharply (\cref{tab:labelspace}): class 3 is absent from
two of four cohorts, and class 2 ranges from 0.15 to 0.44. A per-site operating point therefore
has real headroom, and an oracle-site-label arm gains $+0.0221$ in cross-validation. Every
deployable version we built lost. Site clustering in the embedding space reaches only 0.81--0.84
purity, while the features that cluster sites at 1.00 purity are precisely the hand-crafted
features that are unusable as severity inputs. The mechanism is real and the estimator for it is
not.

\subsection{A possible irreducible component}

The same gait may receive different UPDRS ratings across raters and protocols. If cohort-specific
label calibration is present, part of unseen-site generalization is not recoverable from the
source labels at all, and no amount of representation work would close it. We could not test
this without rater-level metadata. We flag it as the most likely explanation for the residual we
could not move, and note that rater-level annotations would falsify or confirm it directly.

\subsection{Scope}

Everything above is measured on one benchmark, one metric and one hidden test. The claim we
defend is that on this task the released grouping and a first-moment transductive correction
dominate the representation, not that this holds for clinical gait assessment in general. The
negative results bound the families we ran, not the space of methods: they say that no encoder,
alignment, ensemble or decision rule we tested converts, and a different family could.

\section{Reproducibility and data availability}

\paragraph{Code.} The code accompanying this report is public and MIT-licensed at
\url{https://github.com/jlshen025/codabench/tree/main/mocha}. It carries the inference code of
the ranked entry, the 14\,KB trained head that is the system's only trained parameters, and the
three-step chain that produced it. It also carries a verification script. That script assembles
the exact runtime layout, checks every binary's md5, and runs \texttt{predict()} on a probe
built from the \emph{released} cohorts, so the check needs no hidden data and reads no
label. Running it
returns the deployed head's class marginal $[112, 92, 74, 6]$, the marginal recorded for the
ranked entry. The three third-party binaries the entry loads, namely the pretrained encoder, the
SMPL body model and the joint regressor, are referenced with checksums rather than republished;
all three ship with the CARE-PD release.

\paragraph{Data.} All training and evaluation data is CARE-PD~\cite{adeli2025carepd}, used under
CC BY-NC 4.0 and obtained from the official release. The model was trained on the four
UPDRS-labeled cohorts: 3DGait~\cite{wang2023video}, BMCLab~\cite{shida2023public},
T-SDU-PD~\cite{adeli2025carepd} and
PD-GaM~\cite{dadashzadeh2024pecop,adeli2025gaitgen}. The five
unlabeled cohorts named in \cref{sec:task} were read only by the semi-supervised arm reported as
a negative result. The frozen encoder is MotionAGFormer-S~\cite{motionagformer} as released by
its authors, pretrained on Human3.6M~\cite{h36m} and not further trained here. The hidden test
labels were never accessible to us.

\section{Conclusion}

The winning margin on this benchmark did not come from a better motion representation. It came
from taking the reference recipe literally, from aggregating over a grouping that sat in the
input format all along, and from two label-free transductive corrections, on a frozen public
encoder with 14\,KB of trained parameters. For the next entrant, that ordering is the practical
message: read the input format before the model zoo, and price a benchmark's exact training
recipe as part of the benchmark. For the next iteration of the challenge, the ceiling analysis
in \cref{sec:limits} suggests the benchmark is now measuring subject-level severity rather than
walk-level severity, and that scoring subjects explicitly, or releasing rater metadata, would
sharpen what it asks.

The methodological finding we would most want carried forward is the one that cost us the most.
We spent weeks trusting a local cross-validation that was anti-correlated with the score
deciding the competition. A handful of paid reads spent early on calibrating that estimator
against the objective would have paid for themselves many times over. Where a competition
exposes a deterministic scorer on a fixed test set, calibrating the local estimator against it
is not overhead. It is the first experiment.

\section*{Acknowledgements}
We thank the MoCha 2026 organizers for the benchmark, and the CARE-PD team for the dataset,
which is used here under CC BY-NC 4.0. The motion encoder is used frozen and unmodified.
Compute was provided by the Digital Research Alliance of Canada. The author declares no
competing interests.

\bibliographystyle{splncs04}
\bibliography{references}

\end{document}